\pdfoutput=1

\documentclass[11pt]{article}

\usepackage[final]{acl}

\usepackage{times}
\usepackage{latexsym}
\usepackage{wrapfig}
\usepackage{graphicx}
\usepackage{subcaption}
\usepackage{enumitem}

\usepackage[T1]{fontenc}

\usepackage[utf8]{inputenc}

\usepackage{microtype}

\usepackage{inconsolata}

\usepackage{graphicx}

\title{Dipper: \underline{Di}versity in \underline{P}rompts for \underline{P}roducing Large Language Model \underline{E}nsembles in \underline{R}easoning Tasks}

\author{Wenyang Hu$^{* 1,2}$, Gregory Kang Ruey Lau\thanks{Equal contribution.} $^{1,3}$,
Diwen Liu$^{1}$, Jizhuo Chen$^{1}$,\\
\textbf{See-Kiong Ng$^{1}$, Bryan Kian Hsiang Low$^{1}$}\\ 
$^1$National University of Singapore, $^2$SAP,\\
$^3$CNRS@CREATE, 1 Create Way, \#08-01 Create Tower, Singapore 138602\\
\texttt{\{wenyang,greglau,lowkh\}@comp.nus.edu.sg, {seekiong}@nus.edu.sg}\\
}

\usepackage{amsmath,amsfonts,bm, mathtools}

\def\eqref#1{equation~\ref{#1}}

\def\1{\bm{1}}

\DeclareMathAlphabet{\mathsfit}{\encodingdefault}{\sfdefault}{m}{sl}
\SetMathAlphabet{\mathsfit}{bold}{\encodingdefault}{\sfdefault}{bx}{n}

\DeclareMathOperator*{\argmax}{arg\,max}

\DeclareMathOperator*{\diag}{diag}

\usepackage{amsmath,amssymb,amsfonts,bm}
\usepackage{algorithmic, algorithm}
\usepackage[capitalize]{cleveref}
\usepackage{tcolorbox}
\usepackage{xspace}
\usepackage{multirow}
\usepackage{adjustbox}

\newcommand{\algname}{\textsc{Dipper}\xspace}

\newcommand{\subtaskspace}{\mathcal{T}}
\newcommand{\task}{t}
\newcommand{\query}{q}
\newcommand{\fans}{c}
\newcommand{\fansgt}{c^{*}}
\newcommand{\ans}{y}

\newcommand{\model}{M}

\newcommand{\prompt}{w}
\newcommand{\promptspace}{\mathcal{W}}
\newcommand{\ensemble}{\mathcal{E}}
\newcommand{\semspace}{\mathcal{S}}
\newcommand{\semrole}{s}
\newcommand{\valperf}{u}
\newcommand{\vol}{V}

\newcommand{\agg}{\mathcal{A}}

\begin{document}
\maketitle
\begin{abstract}
Large Language Models (LLMs), particularly smaller variants, still struggle with complex reasoning tasks. While inference-time prompting can guide reasoning, existing methods often rely on sequential queries. Ensemble approaches offer a promising path to performance gains, especially given recent batch inference speed-ups. This work introduces \algname, a novel, training-free framework that transforms a single LLM into an effective inference-time ensemble. By feeding the model an optimized and diverse set of prompts in parallel, \algname elicits varied reasoning paths, leading to performance gains. We empirically demonstrate significant improvements on reasoning benchmarks, such as MATH, where a \algname ensemble of three Qwen2-MATH-1.5B instances (via parallel prompting of a single model) outperforms a larger 7B model.

\end{abstract}

\section{Introduction}

Despite remarkable advancements, Large Language Models (LLMs), particularly smaller models that are often constrained by resource limitations (e.g., GPU memory), continue to struggle with complex reasoning tasks \citep{huangReasoningLargeLanguage2023}. While inference-time methods offers a promising approach for enhancing LLM performance, especially for these smaller models \citep{snellScalingLLMTestTime2024}, existing methods like Chain-of-Thought (CoT) and Reflexion often rely on \emph{sequential} LLM queries, thereby incurring additional latency costs~\citep{qiaoReasoningLanguageModel2023, zhengProgressiveHintPromptingImproves2023,yaoTreeThoughtsDeliberate2023}.

Ensemble methods, which involve the use of multiple constituent models in \emph{parallel}, have been shown to improve models' performance and robustness in classical machine-learning settings \citep{ganaieEnsembleDeepLearning2022a} and are promising approaches to achieve better inference-time performance, although less well-studied in the LLM setting. The prospects of applying such methods to LLMs are increasingly attractive, given recent developments that have enabled significant speed-ups in parallel, LLM batch inference. These include methods to efficiently handle key-value cache memory \citep{kwonEfficientMemoryManagement2023} and prompt caching to efficiently reuse common prompts for multiple queries \citep{zhuEfficientPromptCaching2024, gimPromptCacheModular2024}, enabling sub-linear (in the number of queries) costs for batch inference. However, a key challenge for successful ensembles is the \textbf{diversity} among their constituents \citep{kroghNeuralNetworkEnsembles1994,zaidiNeuralEnsembleSearch2020}. This principle extends to LLM ensembles, where achieving meaningful diversity from a single base model remains a central challenge.

Current approaches injecting such diversity, such as using heterogeneous model types (i.e., different LLMs) \citep{jiangLLMBlenderEnsemblingLarge2023,huangEnsembleLearningHeterogeneous2024}, are often impractical due to memory constraints or use preferences for a single model type. Alternatively, methods like self-consistency, which rely on stochastic sampling from the same prompt \citep{wang2023selfconsistency}, typically yield limited diversity, thereby capping potential performance gains. We identify an influential yet overlooked source of diversity: \textbf{system prompts}. LLMs can generate varied reasoning pathways and outputs for the same task when guided by different instructional prompts \citep{kojimaLargeLanguageModels2023}. This observation motivates our central research question: \emph{How can we systematically leverage prompt diversity to construct high-performing LLM ensembles from a single base model efficiently, without model retraining?}

To address this, we introduce \algname, a novel, training-free LLM ensemble framework that constructs an LLM ensemble by feeding a single base LLM an optimized, diverse set of reasoning prompts in parallel. This approach harnesses the parallel processing capabilities of modern LLM inference systems to achieve significant performance improvements in reasoning tasks, particularly for resource-constrained models. \algname is notably simple, resource-efficient, and readily applicable to any black-box LLM via API access. Our key contributions are summarized as follows:

\begin{itemize}[leftmargin=*]
    \item We propose \algname, a novel framework for constructing inference-time ensembles from a single LLM using diverse reasoning prompts, adaptable to any (including black-box) LLM, and detail its core design principles (Sec.~\ref{sec:method}).
    \item We develop a training-free, theory-inspired prompt diversity measure that, when used with our framework, can be efficiently optimized to maximize ensemble performance (Sec.~\ref{sec:diversity}).
    \item We empirically demonstrate that our framework produces significant performance gains on math reasoning tasks (MATH, GSM8K, and MMLU-STEM), where our ensemble consisting of just a few small models (e.g., three Qwen2-MATH-1.5B) can outperform a larger model (e.g., Qwen2-MATH-7B) (Sec.~\ref{sec:exp}).  
\end{itemize}

\section{Background and related works}

\paragraph{LLMs and prompts.} Consider an LLM $\model$ which can be viewed as a black box that encodes conditional probability distribution of text responses $\ans$ over any text input $\query$ and prompt $\prompt$, from which we can sample response $\hat{y}$, i.e.
\begin{equation}
\label{eq:llm_prob}
    \hat{\ans} \sim \model(\query,\prompt) = p_\model(\ans|\query,\prompt).
\end{equation}

In practice, $w$ can be reasoning prompts that instruct LLMs to reason about $q$, e.g., "Let's think step by step" in CoT~\citep{weiChainofThoughtPromptingElicits2023}. These prompts aim to influence the final LLM response, potentially by inducing additional LLM output (e.g., reasoning steps), and have been shown to yield performance boosts.

\paragraph{Prompt optimization. } To alleviate manual effort of prompt engineering, prompt optimization~\cite{zhou2023large, lin2023use, yang2024large, hu2024localized} works aim to automatically search for optimal prompts to maximize an LLM's performance on specific tasks. However, such works have mainly focused on finding the best prompt for a single LLM, leaving the potential of optimizing prompts for LLM ensembles underexplored. In contrast, our work proposes a broader, novel framework for designing inference-time LLM ensembles with diverse prompts, and can incorporate existing prompt optimization methods. For example, we showed that an optimized prompt (e.g., "self-reflection" \citep{shinn2024reflexion}) can be combined with our framework.

\paragraph{LLM ensembles.} Ensemble methods, which combine multiple models to achieve superior performance and robustness \citep{ganaieEnsembleDeepLearning2022a}, have seen limited application in the LLM domain. Prior LLM ensemble works have focused on heterogeneous ensembles that combine outputs from different LLM architectures or API providers \citep{jiangLLMBlenderEnsemblingLarge2023}, multi-agent LLM settings that focus on interactions among agents \citep{duImprovingFactualityReasoning2023,liuDynamicLLMAgentNetwork2023, chenAgentVerseFacilitatingMultiAgent2023}, or homogeneous self-ensembles that generate multiple responses from a single LLM using stochastic sampling \citep{wang2023selfconsistency}. 

However, to the best of our knowledge, we are not aware of any work that has proposed forming and optimizing homogeneous LLM ensembles where their \emph{diversity is injected through varying reasoning prompts} to constituents with the same underlying LLM model. Our work's focus on such an approach exploits LLMs' unique capabilities of generating diverse output given only changes to their prompts, allowing for a simple but effective method to achieve significant training-free boosts to LLM reasoning performance using inference-time compute, especially given recent developments in LLM batch inference methods \citep{kwonEfficientMemoryManagement2023,zhuEfficientPromptCaching2024, gimPromptCacheModular2024}.

\section{Problem formulation} \label{sec:prob_form}

Consider a task $\subtaskspace$ with instances described as tuples $\task \coloneq (\query_t,\fansgt_t)$, where $\query_t$ is a text query and $\fansgt_t$ is the corresponding solution. We denote the response from a single LLM $M$ as $\hat{y} \coloneq \{\hat{r},\hat{\fans}\}$ which consists of its reasoning $\hat{r}$ and final answer $\hat{\fans}$.
We evaluate the performance of the model with a specific prompt $\prompt$, denoted as $M(\cdot,\prompt)$, on the task by computing its expected accuracy over the set of task instances $\subtaskspace$, i.e., $F(M(\cdot, \prompt);\subtaskspace) \coloneq \mathbf{E}_{t \sim \subtaskspace} [\mathbb{I}\{\hat{\fans}_\task = \fansgt_\task\} ]$, which in practice is computed over a representative test set. 

We denote a homogeneous LLM ensemble as $\ensemble(\cdot\,; M,n, \phi)$, consisting of $n$ instances of the same model $M$ and in general has an adjustable inference-time design parameter $\phi$. The ensemble produces a final answer when provided a task query, i.e., $\ensemble(\query_t;M,n,\phi) \rightarrow \hat{\fans}_t$, and we can evaluate its performance based on its expected accuracy:
 \begin{equation}
    \label{eq:objective}
    F(\ensemble,\subtaskspace) = \mathbf{E}_{t \sim \subtaskspace} [\mathbb{I}\{\ensemble(\query_\task;M,n,\phi) = \fansgt_\task\} ].
\end{equation}
Our objective is to design an ensemble framework with an appropriate design parameter $\phi$ such that given fixed $M$, $n$ and a small labeled development set, we can efficiently maximize \cref{eq:objective} by optimizing for $\phi$ to produce the best performing ensemble without additional training.

\section{Method}
\label{sec:method}
A key driver of the ensemble's performance is the diversity present among the constituents in the ensemble. 
Intuitively, a group where every member thinks the same way as each other is likely to result in less robust reasoning and decision-making (e.g. ``groupthink'') compared to a group consisting of members with diverse thinking styles. Similarly, having an ensemble of LLM instances where \emph{all constituents are identical} may be expected to yield less performance advantage compared to a diversified ensemble. In our setting where only a single LLM model is available, self-ensembles \citep{wang2023selfconsistency} are examples of the former, as the constituents rely on LLM sampling stochasticity to generate potentially diverse responses, but will nonetheless still be sampling from the same distribution in \cref{eq:llm_prob} and hence face limited diversity. Our framework for LLM ensembles, \algname, efficiently introduces such diversity at inference time even when only one LLM model is available, through the use of high fidelity, diverse prompts.

In this section, we first provide an overview of our framework \algname, before elaborating on the various components.

\subsection{Overview of the \algname framework}

Drawing inspiration from how using different prompts $\prompt$ would result in varying response distributions in \cref{eq:llm_prob} given the same model $M$, our \algname framework has the set of prompts $\{\prompt_i\}_{i=1}^n$ fed into the ensemble of $n$ LLM instances as the key ensemble design parameter $\phi$.

\algname consists of the following three components: 

\begin{enumerate}[leftmargin=*]
    \item \textbf{Prompt Generator.} First, an LLM generates a large candidate pool of prompts (denoted as $\promptspace$), which can be based on some description of the task and in-context prompt examples that we think may be effective, if such prior knowledge is available. The goal is for the prompts to invoke various types of reasoning pathways when addressing queries, hence injecting diversity into the ensemble. 
    \item \textbf{Prompt Selector.} Drawing parallel to data/prompt selection~\cite{wu2024prompt, wang2025nice, chen2025duet, pinnacle,pied}, we select a subset of $n$ prompts $\{w_i \in \promptspace\}_{i=1}^n$ from the candidate pool of prompts $\promptspace$, where the selection is optimized based on a diversity metric that acts as an approximation of the relative performance of each subset. 
    \item \textbf{Response Aggregator.} Finally, the responses from the $n$ constituent LLMs are aggregated through a response aggregator operation $\agg$ to produce a single final response for the ensemble.
\end{enumerate}

Putting everything together, our \algname framework characterizes an ensemble of size $n$ via $\ensemble(\query_t;M,n,\{\prompt_i\}_{i=1}^n) \coloneq \agg(\{M(q_t,w_i)\}_{i=1}^n) \rightarrow \hat{\fans}_t$, where the subset of prompts $\{\prompt_i\}_{i=1}^n$ is chosen from a candidate pool $\promptspace$ to optimize the expected ensemble performance $F(\ensemble,\subtaskspace)$ for a target task $\subtaskspace$. We now describe each component in detail.

\subsection{Prompt Generator}

The first component plays the important role of generating a large pool of candidate prompts with the following desiderata: 
\begin{enumerate}[leftmargin=*]
    \item \textbf{Fidelity.} Each prompt should be able to influence the LLM into applying a certain type of reasoning approach to the task without significantly degrading the task performance.
    \item \textbf{Diversity.} The prompts should differ sufficiently to elicit various reasoning pathways and provide a diverse selection pool for the subsequent component.
\end{enumerate}

We first show that LLMs are capable of generating prompts that meet these desiderata, via the most direct way of prompting it to generate a pool of candidate prompts while providing it with exemplars illustrating different reasoning prompts. To do so, we considered a list of 7 reasoning prompts inspired by existing works \citep{wangSelfConsistencyImprovesChain2023, deng2023rephrase, yao2022react} on prompting methods to boost reasoning capabilities. Given these prompts as exemplars, we used GPT-4o to further generate a set of 200 different candidate prompts that each represent a different reasoning approach (details in Appx.~\ref{appx:sec:7prompts}). 

\begin{figure}[t]
    \centering
    \resizebox{0.7\columnwidth}{!}{
    \centering\includegraphics{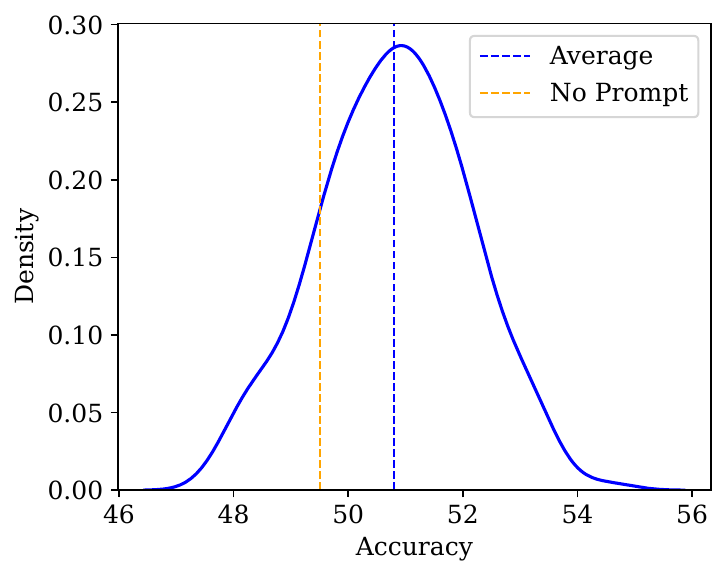}
    }
    \caption{
    The accuracy distribution of 200 candidate prompts on MATH with Qwen2-MATH-1.5B.}
    \label{fig:pool_dist}
\end{figure}

\cref{fig:pool_dist} shows the distribution of average accuracy over a sampled test set of MATH~\citep{hendrycks2021measuring} questions for each prompt, when used with the Qwen2-MATH-1.5B model (i.e., $F(M(\cdot,w);\subtaskspace)$ for $w_i \in \promptspace$). Note that the distribution of accuracy is largely higher than that of the base model without prompts, and similar to the accuracies achieved by the reasoning prompt exemplars, demonstrating the fidelity requirement. Qualitatively, we see that the prompts are also relatively diverse -- they generally specify certain reasoning approaches inspired by various subject domains (see Appx.~\ref{app:sec:200prompts}). We will quantify this diversity in Sec.~\ref{sec:diversity} with our proposed metric.

Note that when generating the prompts, we did not pass any task description to the LLM prompt generator. We did so as the reasoning prompts can be task-agnostic. In practice, the candidate pool of reasoning prompts need not be generated on-the-fly, but can be drawn from a shared pool prepared beforehand by a more powerful LLM, to be used by ensembles consisting of much smaller LLMs, as we demonstrated. The actual selection of relevant prompts from this larger pool can then be done by the prompt selector component, which we will describe next in Sec.~\ref{sec:diversity}.

\subsection{Prompt Selector} \label{sec:diversity}

With our framework, the optimization problem in \cref{eq:objective} reduces to an optimization to choose the best subset of prompts $\{\prompt_i\}_{i=1}^n$ from the set of candidate prompts $\promptspace$:
\begin{equation}
    \label{eq:div_objective}
    \argmax_{\{w_i \in \promptspace \}_{i=1}^n} F(\ensemble(\query_\task;M,n,\{w_i\}_{i=1}^n), \subtaskspace).
\end{equation}
Unfortunately, directly optimizing \cref{eq:objective} is a combinatorial problem that is very challenging, even if a development/validation set is available for the task of interest. 
For example, selecting 5 prompts from a candidate pool of 200 prompts involves searching over ${200 \choose 5} \approx 2.5 \times 10^9$ candidates. 
Instead, we note that the best ensemble composition requires a balance of the two desiderata: fidelity and diversity.
Hence, we propose optimizing \cref{eq:objective} by considering how to prioritize the prompts that have the best predicted performance on the task $\subtaskspace$, while maximizing the diversity of the selected set of prompts. Our method draws inspiration from past works on determinantal point processes (DPP) \citep{kuleszaDeterminantalPointProcesses2012, lau2025uncertainty}, which consider similarity kernels comprising separate quality and diversity terms that match our requirements.

\paragraph{Prompt fidelity.}
First, we can approximate the predicted performance of each prompt by its average performance on a task development set $\subtaskspace_d$\footnote{Without such a development set, an uninformed prior on the performance (e.g. uniform distribution across roles), or an informed-prior based on domain knowledge, could also be used.}. Note that as inference using these various prompts on a small development set can be done in parallel, this process can in practice be significantly sped up by existing batch inference techniques such as those employed by vLLM \citep{kwonEfficientMemoryManagement2023}.

Specifically, for a candidate pool of prompts $\promptspace$ and development set $\subtaskspace_d$, we can define a prompt fidelity mapping $\valperf:\promptspace \rightarrow [0,1]$,
\begin{equation}
\label{eq:valperf}
    \valperf(\prompt) \coloneq F(M(\cdot,\prompt),\subtaskspace_d), 
\end{equation}
where $M(\cdot,\prompt)$ is the LLM model conditioned by prompt $\prompt \in \promptspace$, and $F$ the expected accuracy defined in \cref{sec:prob_form}. In practice, for a candidate pool of size $n$, $\valperf(\prompt)$ can be represented as an $n \times 1$ column vector, with the elements representing each prompt's expected accuracy.

\paragraph{Semantic entropy.} 

Then, we measure prompt diversity by considering how different the semantic meanings of the $n$ role prompts are from each other. We represent each prompt's semantic meaning with a mapping $R$ from its text representation $\prompt$ into a normalized continuous vector $\semrole \in \mathbb{R}^{p}$ in a $p$-dimensional semantic embedding space $\semspace$ through a sentence embedding model $M_s$ \citep{reimers-2019-sentence-bert}, i.e., $R(w)\coloneq M_s(w)$. This mapping can be represented as an $n\times p$ prompt embedding matrix $R = [\semrole_1,\cdots,\semrole_n]$ where $\semrole$ is a $1\times p$ row vector representing each prompt. 

To quantify prompt diversity of a given set of prompts, we propose to compute the volume enclosed by the selected prompts in semantic space. Intuitively, for $n$ fixed prompts, more diverse prompts point to more varied directions in semantic space, and enclose a larger volume. Specifically, we note that from basic geometry, the determinant of a Gram matrix is the squared volume of the parallelepiped spanned by the embedding vectors. Hence, we define the semantic volume metric $\vol$ as 
\begin{equation}
\label{eq:sem_vol}
    \vol \coloneq \log \det (RR^T),
\end{equation}
where we take the logarithm (for numerical stability) of the Gram matrix determinant\footnote{We omit a factor of 2 which does not affect the optimization process. For our setting, we also have $n<p$ as the semantic embedding space is usually high-dimensional.} 
Sec.~\ref{app:sec:illu_logdet} shows how sets of prompts that are qualitatively observed to be more diverse have larger quantitative semantic volume. 

\paragraph{Fidelity-adjusted semantic volume (FASV).}
To incorporate the prompts' expected accuracy information, we can compute the performance-adjusted prompt embedding matrix, 
\begin{equation}
\label{eq:tilde_R}
  \Tilde{R}\coloneq \exp({\frac{\alpha}{2} \diag(u)})R, 
\end{equation}
where $\diag(u)$ is the diagonal matrix with its $i^\text{th}$ diagonal element being the corresponding element $u_i$. This essentially scales each row $\semrole_i$ in $R$ by an exponential factor based on its corresponding predicted accuracy, $\exp({\frac{\alpha}{2} u_i})$, where $\alpha$ is a scalar hyperparameter influencing the balance between diversity and expected performance. Intuitively, prompts with higher expected accuracy would then be able to support larger semantic volume and hence be prioritized for inclusion into the ensemble. The adjusted embedding matrix can then be used to compute the semantic volume in \cref{eq:sem_vol}, which simplifies to
\begin{equation}
\label{eq:tilde_vol}
  \tilde{\vol} = \log \det (\tilde{R}\tilde{R}^T) = \vol + \alpha \|\valperf\|_{1} \ , 
\end{equation}
providing an interpretable expression illustrating the balance between the diversity (i.e., the semantic volume metric in \cref{eq:sem_vol}) and fidelity desiderata (i.e., the L1 norm of the prompt fidelity metric in (\cref{eq:valperf}) that needs to be optimized for the ensemble. Derivation details are in \cref{app:sem_vol_derive}, and we provide empirical analysis of the effectiveness of this combined metric in \cref{sec:fid-div}.

\begin{algorithm}[htb]
    \caption{\algname FASV algorithm}\label{alg:div}
    \begin{algorithmic}[1]
    \STATE \textbf{Input:} LLM model $M$, Initial candidate prompt set $\bar{\promptspace}$, Semantic embedding model $M_s$, Development set $\subtaskspace_d$, Ensemble size $n$, Fidelity-diversity hyperparam $\alpha$
    \STATE \textbf{Output}: Ensemble prompt set $\mathcal{Z}$
    \STATE $\mathcal{Z} \gets \{\ \}$
    \STATE $\bar{\valperf}(w) \gets [F(M(\cdot,\prompt_i),\subtaskspace_d) \text{ for } w_i \in \bar{\promptspace}]$ 
    \STATE $\mathcal{Z} \gets \mathcal{Z} \cup \arg\max_{w} \bar{u}(w)$
    \STATE $\promptspace \gets \bar{\promptspace} \setminus \arg\max_{w} \bar{u}(w)$
    \FOR {$j = 1,\ldots, n$}
        \STATE $\Tilde{\mathcal{V}} \gets [\ ]$
        \FOR {$w_k \in \promptspace$}
        \STATE $\mathcal{P} \gets \mathcal{Z} \cup w_k$
            \STATE $u(w) \gets [F(M(\cdot,\prompt_i),\subtaskspace_d) \text{ for } w_i \in \mathcal{P}]$
            \STATE $R(w) \gets [M_s(w_i) \text{ for } w_i \in \mathcal{P}]$
            \STATE $\Tilde{\vol}_{w_k}  \gets \log \det (RR^T) + \alpha \|\valperf\|_{1}$
            \STATE $\Tilde{\mathcal{V}}(w) \gets [ \Tilde{\mathcal{V}}(w),\Tilde{\vol}_{w_k}]$
            \ENDFOR
        \STATE $\mathcal{Z} \gets \mathcal{Z} \cup \arg\max_{w} \Tilde{\mathcal{V}}(w)$
        \STATE $\promptspace \gets \promptspace \setminus \arg\max_{w} \Tilde{\mathcal{V}}(w)$
        \ENDFOR
    \RETURN $\mathcal{Z}$
    \end{algorithmic}
\end{algorithm}

\paragraph{Optimization of semantic entropy.}
We can now recast \cref{eq:objective} as an optimization of the fidelity-adjusted semantic volume metric $\tilde{\vol}$ evaluated over the set of candidate prompts. Note that instead of the expected ensemble performance $F(\ensemble)$, which is an objective that can only be 
optimized by blackbox optimization methods like Bayesian Optimization \citep{BObook,qbo, readme}, our metric $\tilde{\vol}$ can be efficiently approximated by well-established heuristics. 

Specifically, as the semantic volume metric is submodular, we can optimize for the best subset of roles by incrementally building the subset with a greedy approach up to the desired size $n$ and still be guaranteed a good approximation \citep{submodular}. This is an important advantage that allows us an efficient and theoretically-inspired approach to obtain the best ensemble prompts. Our proposed algorithm is outlined in \cref{alg:div}.

\subsection{Response Aggregator}\label{sec:agg}

Given the responses from the various LLMs of constituents, the aggregation method determines how much information from the constituents is used to derive the final output of the ensemble. We consider the two most popular approaches:

\paragraph{Majority voting (MV)} It involves extracting the final answer $\hat{c}$ from each LLM response $\hat{y}=\{\hat{r},\hat{c}\}$, and selecting the answer that has been proposed the most number of times. This approach does not evaluate the quality of reasoning $\hat{r}$ output produced by each LLM, but is easily implementable.

\paragraph{Best-of-N} An external reward model is implemented to evaluate the response $\hat{y}$ of each agent, and the response with the highest score is selected as the final response. This approach does not leverage consensus among constituents but could be effective in identifying the correct responses that would be only covered by a few agents.

\section{Experiments}
\label{sec:exp}

\textbf{Experimental set-up.} We empirically evaluate our framework on mathematically reasoning tasks with the MATH~\citep{hendrycks2021measuring}, GSM8K, and MMLU-STEM datasets. We implement our framework using the GPT-4o as our prompt generator and Qwen2-MATH-1.5B as the constituent model in the ensemble, where the ensemble constituents are run in parallel using vLLM \citep{kwonEfficientMemoryManagement2023} for fast batch inference. Further details (Appx.~\ref{appx:exp:detail}) and additional results (Appx.~\ref{appx:sec:more_results}) are in the Appendix.

\paragraph{Baselines.} We evaluate our \algname framework by comparing it against the "Self-ensemble" method, which lacks prompt diversity but incorporates diversity through repeated response sampling \citep{wang2023selfconsistency}, along with the single model performance as a reference. We also include two other implementation variants of \algname in our analysis, beyond the implementation based on semantic volume, "Dipper (FASV)":
\begin{enumerate}
    \item \textbf{Random+.} Here we randomly sample prompts from the candidate pool based on a probability distribution proportional to their predicted accuracy as defined in \cref{eq:valperf}, i.e., $p(w) \propto \valperf(w)$. This aims to achieve diversity through the sampling process while prioritizing prompts with higher predicted accuracy.
    \item \textbf{Top-n.} Here we greedily select the top $n$ prompts which are ranked based on their predicted accuracy $\valperf(w)$. It assumes that the diversity of prompts introduced by our prompt generation process is sufficient and hence does not explicitly optimize for ensemble diversity during the prompt selection phase. 
\end{enumerate}

\begin{figure}[t]
    \centering
    \resizebox{0.8\columnwidth}{!}{
    \centering\includegraphics{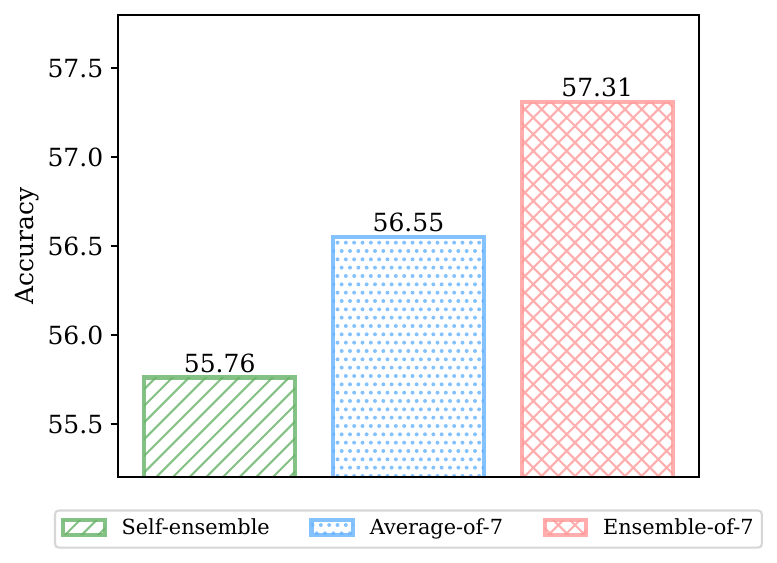}
    }
    \caption{Comparison of different ensembles of 7 reasoning prompts on MATH.}
        \label{fig:7prompts:acc}
\end{figure}

\begin{figure}[t]
    \centering
    \resizebox{0.8\columnwidth}{!}{
    \centering\includegraphics{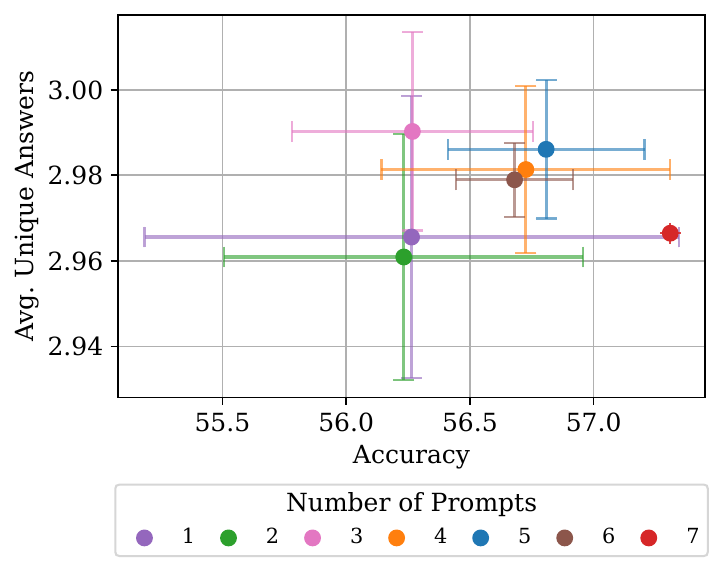}
    }
    \caption{Accuracy vs. average number of unique answers using different numbers of prompts in ensembles.}
        \label{fig:7prompts:div}
\end{figure}

\subsection{Ensembles with fixed prompt methods}\label{sec:7prompts}

To motivate our \algname framework and demonstrate the importance of prompt diversity, we first consider a fixed set of seven distinct reasoning prompts inspired by existing works \citep{wangSelfConsistencyImprovesChain2023, deng2023rephrase, yao2022react} (details in Appx.~\ref{appx:sec:7prompts}). With a fixed ensemble size of seven, \cref{fig:7prompts:acc} shows that an ensemble using these seven different prompts (57.31\%) outperforms both a baseline self-ensemble without prompt variation (55.76\%) and the average performance (56.55\%) of seven self-ensembles, each using only one of the distinct prompts.

In addition, we evaluated the impact of prompt diversity by constructing ensembles with varying numbers of unique prompts (from one to seven) drawn from this set, while maintaining an ensemble size of seven. When fewer than seven unique prompts were used, responses were randomly sampled to meet the ensemble size. The result in \cref{fig:7prompts:div} indicates that increasing the number of unique prompts generally leads to higher accuracy and reduced variance. This suggests that prompt diversity within an ensemble can enhance performance and consistency, particularly when the performances of prompts are unknown before the final evaluation.

\begin{figure}[H]
\begin{minipage}{0.48\textwidth}
\centering
\includegraphics[width=0.82\textwidth]{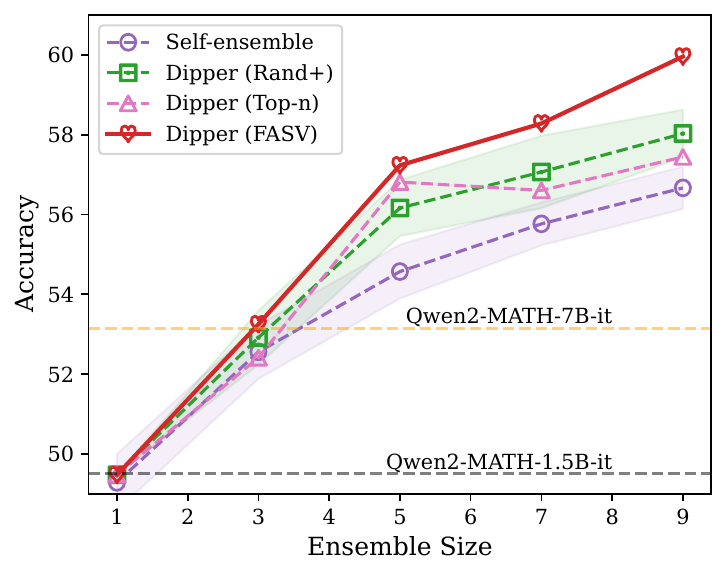}
\captionsetup{width=0.9\textwidth}
\vspace{-3mm}
\caption{Comparison of different ensemble methods on MATH for Qwen2-MATH-1.5B.}
\label{fig:exp:math}
\end{minipage}
\hfill
\begin{minipage}{0.48\textwidth}
\centering
\includegraphics[width=0.82\textwidth]{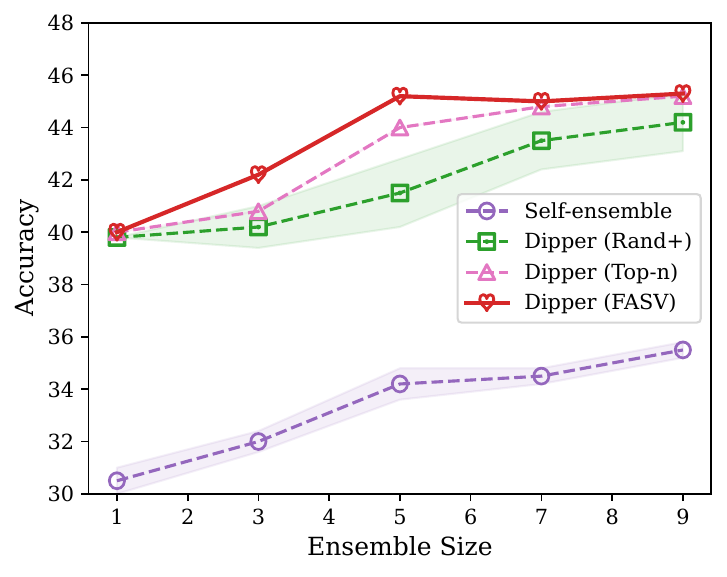}
\captionsetup{width=0.9\textwidth}
\vspace{-3mm}
\caption{Comparison of different ensemble methods on MATH for the LLaMA3.2B model.}
\label{fig:llama_MATH}
\end{minipage}
\end{figure}

\subsection{Ensembles with optimized prompt diversity}
Next, we consider our full \algname framework. We first generate a pool of prompt candidates ($|\promptspace|=200$) using the 7 reasoning prompts in the previous section as in-context exemplars (details in Appx.~\ref{appx:sec:7prompts}) and then perform prompt fidelity-diversity optimization (Sec.~\ref{sec:diversity}) to select the best ensemble prompts. As shown in \cref{fig:exp:math}, our full \algname implementation with FASV achieves the highest accuracy compared to the self-ensemble baseline and all other \algname variants across various ensemble sizes. \algname also significantly outperforms the single LLM. For example, \algname with $n=9$ has close to a 10\%-pt increase (\textasciitilde20\% accuracy gain) compared to the single LLM baseline. In fact, our ensemble that consists of just 3 Qwen2-MATH-1.5B models already (slightly) outperforms the next model size class, the Qwen2-MATH-7B model. Note also that the performance gain of \algname over the self-consistency baseline is about as large as the gain from moving up one model class (from the 1.5B to 7B model). We see similar results on MATH with the general model LLaMA3.2B in Fig.~\ref{fig:llama_MATH}, where \algname (FASV) is shown consistently effective. Refer to \cref{appx:sec:more_results} for more results for other datasets (e.g., GSM8K, MMLU-STEM, BIG-Bench) and models.

\subsection{Fidelity-diversity optimization}
\label{sec:fid-div}

\begin{figure}[t]
    \centering
    \resizebox{0.85\columnwidth}{!}{
    \centering\includegraphics{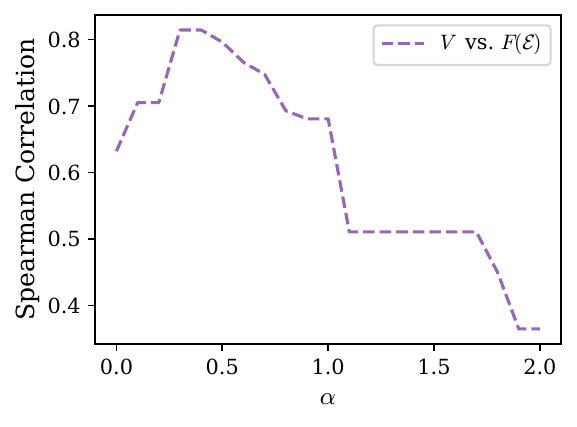}
    }
    \vspace{-4mm}
    \caption{
    Spearman correlation between $\vol$ and test performance $F(\ensemble)$ on the MATH under different fidelity-diversity hyperparameter $\alpha$.
    }
    \label{fig:math_alpha}
    \vspace{-4mm}
\end{figure}

To further understand the mechanisms behind \algname's performance gains, we analyze the predictive power of our fidelity-adjusted semantic volume metric in \cref{eq:tilde_vol} (which we denote as $\vol$ in this section for notational simplicity) on the final ensemble performance on the test set $F(\ensemble)$. We quantify this by computing the Spearman correlation between $V$ and $F(\ensemble)$: the higher the Spearman correlation, the better our optimization of the ensemble prompts via $\vol$ will lead to higher ensemble test performance. \cref{fig:math_alpha} shows the Spearman correlation of $\vol$ and $F(\ensemble)$ for the MATH dataset experiment, with different fidelity-diversity hyperparameter $\alpha$ values. We can observe two key insights.

\begin{figure}[t]
    \centering
    \resizebox{0.85\columnwidth}{!}{
    \centering\includegraphics{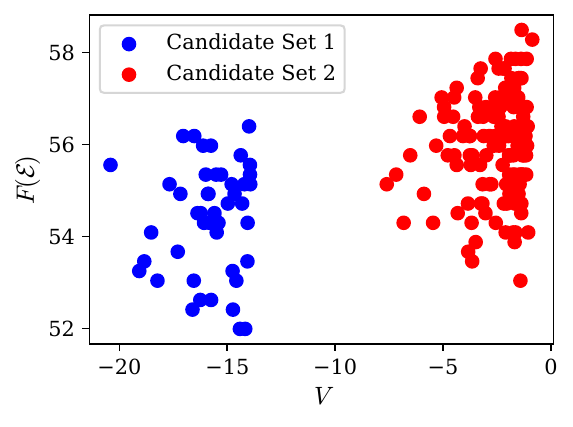}
    }
    \vspace{-4mm}
    \caption{
    The scatter plot showing correlation between $V$ and $F(\ensemble)$ for different prompt candidates.
    }
    \label{fig:para_scatter}
\end{figure}

First, there is a relatively strong positive correlation between $\vol$ and $F(\ensemble)$, going as high as $0.8$ for some values of $\alpha$. This corroborates our main results where our \algname method that explicitly optimizes for $\vol$ outperforms other baselines and achieves higher $F(\ensemble)$. 

Second, there is a U-shape trend between the Spearman correlation and hyperparameter value $\alpha$, where the correlation increases as $\alpha$ increases from 0, but decreases after a certain point. This trend demonstrates the need of taking into account both fidelity and diversity when optimizing for the set of ensemble prompts, as we discussed in \cref{sec:diversity}. On the one hand, $\alpha=0$ corresponds to the case where we focus solely on diversity and ignore the fidelity or individual predicted performance of prompts ($\valperf(w)$) -- this may select a set of diverse prompts, but potentially some irrelevant or poor performing prompts. However, if we emphasize fidelity too much and disregard diversity, we may end up selecting very similar prompts resulting in less ensemble performance gains. At the extreme, choosing large $\alpha$ reduces to the Top-n baseline implementation, which has poorer performance than \algname that optimizes for semantic volume. In practice, just like other machine learning hyperparameters, we could inform the choice of $\alpha$ with the development set, if available.

\subsection{Prompt candidate matters}
We then analyze how the candidate pool diversity introduced by Prompt Generator contributes to our framework. 
Out of the original candidate set $\promptspace$ (Candidate set 2), we obtained another set by selecting one cluster after performing k-means clustering over $\promptspace$ with $k=4$ ($\promptspace'$, Candidate set 1). We then randomly select ensembles of size $n=5$, and plot their respective $\vol$ and $F(\ensemble)$ in \cref{fig:para_scatter}. We can see that ensembles from $\promptspace'$ have much lower accuracy and semantic volume compared to those from $\promptspace$, illustrating the importance of the candidate pool diversity from the Prompt Generator.

\subsection{\algname combined with other prompting methods like Reflexion}

In addition, we also show that our ensemble framework \algname is \textbf{orthogonal} to other established prompting techniques (e.g. CoT and Reflexion \citep{shinn2024reflexion}), allowing it to stack and bring greater performance. To demonstrate this, we first use \algname to select 5 agents and query each agent with questions from the MATH dataset. Their initial responses will then be self-reflected according to the method proposed in Reflexion \citep{shinn2024reflexion}, before being aggregated into the final answer with MV. We found that combining self-reflection with \algname achieves a performance gain of $8\%$ (from an accuracy of $57\%$ to $65\%$), demonstrating that \algname has the potential to be extended further or combined with other methods.

\begin{figure}[t]
    \centering
    \resizebox{0.8\columnwidth}{!}{
    
    \centering\includegraphics{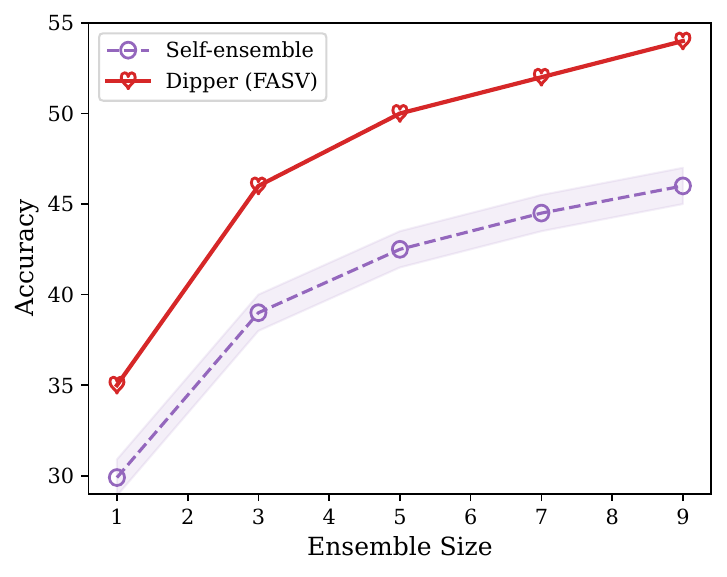}
    }
    \caption{
    Comparison of \algname and self-ensemble baseline on MATH using LLaMA-3B and Best-of-N aggregation.
    }
    \label{fig:best_of_n:math}
\end{figure}

\subsection{Generalization to Best-of-N aggregation}
Finally, we study the effects of using the Best-of-N aggregation for our response aggregator component, showing that \algname can work well with external reward models. We use an existing reward model, ``Qwen2.5-Math-RM-72B'' to assess the quality of the generated responses and select the final response, using the LLaMA-3B model and Best-of-N aggregation in our \algname framework. As seen in \cref{fig:best_of_n:math}, the \algname variants significantly beats the self-ensemble baseline when Best-of-N is used. In this scenario, the performance among the \algname variants are closer since Best-of-N considers each constituent individually rather than jointly (like in MV) to produce the final answer, though our full \algname implementation still consistently performs the best. We also see that \algname can stack with benefits from a strong verifier model, given its performance gains compared to the result in \cref{fig:exp:llama:math} where no verifier model is available.

\section{Conclusion}
In this work, we have proposed a novel framework, \algname, where a single LLM model type is fed an optimized, diverse set of reasoning prompts in parallel, effectively producing an ensemble at inference time to achieve performance improvement in reasoning tasks. Our empirical findings have demonstrated the effectiveness of various \algname implementations in improving inference performance for a variety of reasoning tasks, which may inspire future works to investigate additional optimization methods for prompt-based inference-time ensembles to further improve performance gains.

\clearpage

\section*{Limitations}

Our framework \algname focuses on developing inference-time ensembles where each constituent is based on the same base model -- this caters to the most common and straightforward scenario where users are using a single LLM model and can apply \algname to further boost its reasoning performance at inference time without additional training. However, when users may wish to use heterogeneous models, \algname currently does not take into account such model diversity, which we believe may enable further performance boosts if properly optimized. We leave it to future works to potentially build on \algname to extend it beyond its current limitations in this regard.

\section*{Acknowledgment}
This research/project is supported by the National Research Foundation, Singapore under its AI Singapore Programme (AISG Award No: AISG2-PhD/2023-01- 039J). This research is part of the programme DesCartes and is supported by the National Research Foundation, Prime Minister’s Office, Singapore under its Campus for Research Excellence and Technological Enterprise (CREATE) programme. This research/project is supported by the National Research Foundation, Singapore under its National Large Language Models Funding Initiative (AISG Award No: AISG-NMLP-2024-001). Any opinions, findings and conclusions or recommendations expressed in this material are those of the author(s) and do not reflect the views of National Research Foundation, Singapore. This research/project is also supported by SAP and Singapore’s Economic Development Board under the Industrial Postgraduate Programme.

\bibliography{references, LLMAgents}

\clearpage

\appendix
\onecolumn

\section{Detailed Experimental Setting}\label{appx:exp:detail}

The huggingface model path for the primary model we used is ``Qwen/Qwen2-Math-1.5B-Instruct'' and the sentence transformer's model path is `all-MiniLM-L6-v2''. We use the default generation parameters for Qwen2-Math-1.5B-Instrct: the temperature is set to 0.7, the top probability used for filtering is set to 0.8, and the repetition penalty is set to 1.05. We also set the max tokens to be generated to 512.

\subsection{Fixed 7 prompts and Prompt Generation}\label{appx:sec:7prompts}

We consider 7 prompts inspired by existing works and list them in Tab.~\ref{appx:tab:7prompt} below.

\begin{table}[h!] 
\centering
\caption{The table of 7 basic reasoning prompts inspired by existing works.}
\begin{tabular}{p{0.95\columnwidth}}
\hline
\textbf{Prompt} \\
\hline
Let's think step-by-step to find the answer. \\
\hline
Reflect on the question carefully before answering. \\
\hline
Rephrase the question in your own words before responding. \\
\hline
Actively reason through the question and answer each part systematically. \\
\hline
Answer this question as a scientist would. \\
\hline
Eliminate the obviously incorrect answers first and then choose the most likely correct answer. \\
\hline
Analyze the context of the question and use relevant information to derive the answer. \\
\hline
\end{tabular}
\label{appx:tab:7prompt}
\end{table}

We use the prompt template in Tab.~\ref{appx:tab:generation:propmt} to generate 200 diverse prompts.

\begin{table}[h!]
\centering
\caption{The prompt template for generating more reasoning prompts based on the 7 prompts.}
\begin{tabular}{p{0.95\columnwidth}}
\hline
\textbf{Prompt Generation Template} \\
\hline
Here are some instruction examples: \\
\\
\textcolor{blue}{\{7 reasoning prompts\}} \\
\\
Study the above examples and brainstorm 200 similar instructions with detailed descriptions of\\different reasoning behaviors that are helpful for reasoning. Those 200 proposed instructions \\should be diverse enough. \\

\hline
\end{tabular}
\label{appx:tab:generation:propmt}
\end{table}

\subsection{Evaluation}\label{appx:sec:evaluation}

We primarily consider three datasets in our paper. For MATH, we randomly sample 10\% test samples from each category in its official test split and form a fixed subset of size 500. We then uniformly randomly sample 20 samples from this subset to create a validation dataset and use the rest 480 samples as the hold-out test dataset. For GSM8K and MMLU-STEM, we use their official split of test data and uniformly randomly sample 20 samples to form a validation dataset for each task, and use the rest samples as the hold-out test data.

In the inference evaluation, we use 4-shot exemplars for MATH, 8-shot for GSM8K, and 5-shot for MMLU-STEM. Those exemplars are adopted from the evaluation setting in Qwen2-MATH~\citep{qwen2math} and fixed for all questions and all methods.

\clearpage

\section{Fidelity-adjusted semantic volume metric}
\label{app:sem_vol_derive}

In this section, we provide the explicit derivation of how our fidelity-adjusted semantic volume metric can be simplified to a weighted sum of two terms representing the diversity and fidelity desiderata in \cref{eq:tilde_vol}, which clearly illustrates the balance between the two desiderata during the optimization process.

\begin{align}
    \tilde{\vol} =& \log \det (\tilde{R}\tilde{R}^T) \\
    =& \log \det \left(\exp({\frac{\alpha}{2} \diag(u)})R \right ) \left(\exp({\frac{\alpha}{2} \diag(u)})R \right )^T \label{eq:8} \\
    =& \log \left [ \det \left(\exp({\frac{\alpha}{2} \diag(u)})\right) \det \left(R R^T\right) \det \left(\exp({\frac{\alpha}{2} \diag(u)})^T \right) \right ] \label{eq:9} \\
    =& \log \det (RR^T) + 2 \log \det \left(\exp({\frac{\alpha}{2} \diag(u)}\right) \label{eq:10}\\
    =& \vol + 2 \log \prod_i \exp({\frac{\alpha}{2} u_i}) \label{eq:11}\\
    =& \vol + \alpha \|\valperf\|_{1} \ \label{eq:12}
\end{align}
where \cref{eq:8} follows from the definition of $\Tilde{R}$ in \cref{eq:tilde_R}, \cref{eq:9} uses the identity $\det(AB)=\det(A)\det(B)$, \cref{eq:10} the identity $\log(AB)=\log(A)+\log(B)$, \cref{eq:11} the definition of semantic volume in \cref{eq:sem_vol},
and \cref{eq:12} noting that $\sum_i u_i = \| u \|_1$ since $u \geq 0$.

\section{Additional Results} \label{appx:sec:more_results}

\subsection{Results on General-Purpose Model}

\begin{figure}[ht]
    \centering
    \begin{minipage}[t]{0.45\textwidth}
        \centering
        \includegraphics[width=\textwidth]{figures/llama_MATH.pdf}
        \caption{Comparison of different ensemble methods on MATH for the LLaMA3.2-3B-it model.}
        \label{fig:exp:llama:math}
    \end{minipage}\hfill
    \begin{minipage}[t]{0.45\textwidth}
        \centering
        \includegraphics[width=\textwidth]{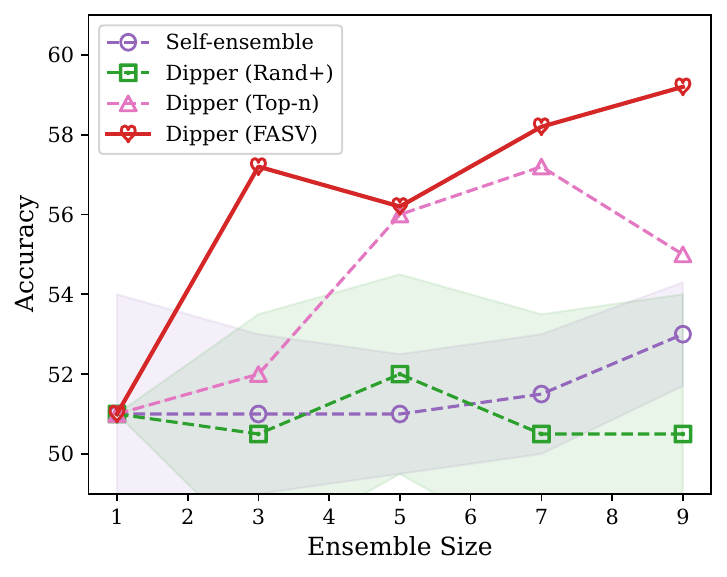}
        \caption{Comparison of different ensemble methods on MMLU-STEM for the LLaMA3.2-3B-it model.}
        \label{fig:exp:llama:mmlu}
    \end{minipage}
\end{figure}

To show our method \algname also generalizes to a general-purpose model (e.g., LLaMA), we evaluate its performance using LLaMA3.2-3B-it model on MATH and MMLU-STEM. The results are presented in \cref{fig:exp:llama:math} and \cref{fig:exp:llama:mmlu}, which demonstrate that our full \algname implementation consistently outperforms the self-ensemble baseline and other \algname variants across datasets.

To demonstrate the generalization of our method \algname to more recent models, we compare its variants against the baseline method on the new Qwen3-0.6B and Qwen3-1.7B models. The results are presented in Tab.~\ref{tab:qwen3:06b} and \ref{tab:qwen3:1:7b}, which suggest that our method \algname still has the same performance advantage on recent LLMs.

\begin{table}[h]
\centering
\caption{Comparison of different ensemble methods on MATH for Qwen3-0.6B}
\label{tab:qwen3:06b}
\begin{tabular}{lcccc}
\hline
Method & n=3 & n=5 & n=7 & n=9 \\
\hline
Self-ensemble & 42.18 & 44.33 & 45.12 & 45.43 \\
Dipper (Rand+) & 42.18 & 45.03 & 46.31 & 46.98 \\
Dipper (Top-n) & \textbf{42.98} & 44.65 & 45.91 & 47.80 \\
Dipper (FASV) & \textbf{42.98} & \textbf{44.86} & \textbf{46.54} & \textbf{49.26} \\
\hline
\end{tabular}
\end{table}

\begin{table}[h]
\centering
\caption{Comparison of different ensemble methods on MATH for Qwen3-1.7B}
\label{tab:qwen3:1:7b}
\begin{tabular}{lcccc}
\hline
Method & n=3 & n=5 & n=7 & n=9 \\
\hline
Self-ensemble & 47.38 & 49.54 & 51.07 & 51.70 \\
Dipper (Rand+) & 50.29 & 51.93 & 52.94 & 53.50 \\
Dipper (Top-n) & 51.36 & \textbf{53.25} & 53.04 & 53.04 \\
Dipper (FASV) & \textbf{51.57} & 52.62 & \textbf{53.25} & \textbf{54.51} \\
\hline
\end{tabular}
\end{table}

\subsection{Results on more datasets for the Qwen2-MATH-1.5B model}
\begin{figure}[ht]
    \centering
        \begin{minipage}[t]{0.45\textwidth}
        \centering
        \includegraphics[width=\textwidth]{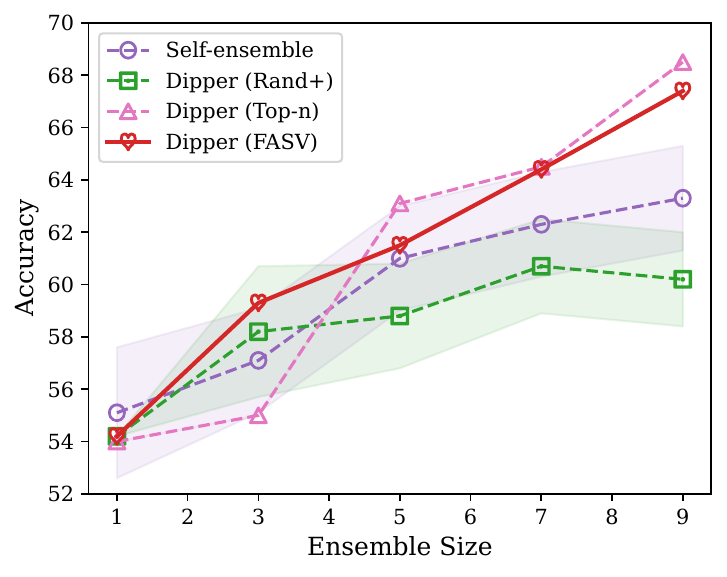}
        \caption{Comparison of different ensemble methods on MMLU-STEM. \algname variants outperform the self-ensemble baseline, consistent with the other experiments.}
        \label{fig:exp:mmlu}
    \end{minipage}
    \hfill
    \begin{minipage}[t]{0.45\textwidth}
        \centering
        \includegraphics[width=\textwidth]{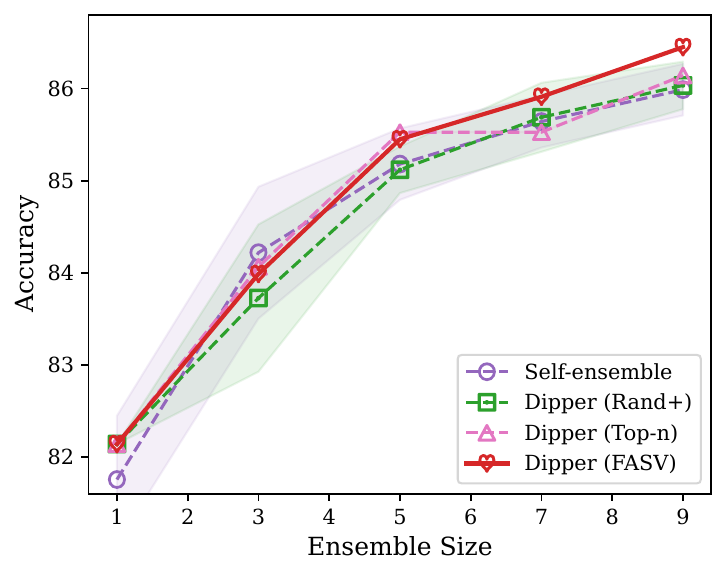}
        \caption{Comparison of different ensemble methods on GSM8K. \algname still outperforms the self-ensemble baseline, although gains are not as obvious as in other benchmarks as it is an easier task where the base model can already perform well.}
        \label{fig:exp:gsm8k}
    \end{minipage}\hfill

\end{figure}

Apart from the MATH dataset, we also evaluate the performance of \algname using the Qwen2-MATH-1.5B model on MMLU-STEM and GSM8K. The results in \cref{fig:exp:mmlu} and \cref{fig:exp:gsm8k} again demonstrate that our full \algname implementation can consistently outperform the self-ensemble baseline and achieve superior or comparable results against the other \algname variants. The performance gains in GSM8K is more limited compared to the gains in experiments for other datasets as it is an easier dataset where the base model can already achieve high accuracy. As can be seen across all our experimental results, our full \algname implementation comprising the theoretically-inspired semantic volume diversity optimization component achieves the most consistent performance, unlike some of the other variants.

\subsection{Results beyond Reasoning Tasks}

To investigate the effectiveness of \algname extending to non-reasoning tasks, we have conducted additional experiments on three challenging BIG-Bench tasks following the Instruction Induction setting from \citet{zhou2022large}. As per our framework, our prompt generator first generates instruction candidates based on 5 randomly sampled demonstrations in the form of input-output pairs, before our prompt selector optimizes for the ensemble prompt composition. As shown in Tab.~\ref{tab:bigbench}, our FASV variant consistently outperforms others, especially over the self-ensemble baseline and when the ensemble size becomes larger. This indicates that \algname has a potential to be deployed as a general inference framework for improved performance.

\begin{table}[h]
\centering
\caption{Performance Comparison on BigBench Tasks using Llama3.1-8B}
\label{tab:bigbench}
\adjustbox{max width=\textwidth}{
\begin{tabular}{l|cccc|cccc|cccc}
\hline
\multirow{2}{*}{Method} & \multicolumn{4}{c|}{synonyms} & \multicolumn{4}{c|}{word\_unscrambling} & \multicolumn{4}{c}{word\_sorting} \\
& n=3 & n=5 & n=7 & n=9 & n=3 & n=5 & n=7 & n=9 & n=3 & n=5 & n=7 & n=9 \\
\hline
Self-ensemble & 4.38 & 5.0 & 5.88 & 4.88 & 10.25 & 13.0 & 15.25 & 17.38 & \textbf{42.125} & \textbf{44.25} & 45.25 & 46.0 \\
Dipper (Rand+) & \textbf{11.5} & 13.5 & 14.25 & 14.75 & \textbf{26.27} & 27.6 & 28.0 & 28.0 & 33.875 & 39.5 & 39.25 & 40.5 \\
Dipper (Top-n) & 10.0 & 13.75 & 12.5 & 17.5 & 22.67 & 24.0 & 25.33 & 25.33 & 28.75 & 28.75 & 36.25 & 33.75 \\
Dipper (FASV) & 6.25 & \textbf{17.5} & \textbf{18.75} & \textbf{18.75} & 25.33 & \textbf{28.0} & \textbf{29.33} & \textbf{33.33} & 38.75 & 43.75 & \textbf{47.5} & \textbf{50.0} \\
\hline
\end{tabular}
}
\end{table}

\subsection{More Results on Prompt Diversity}

\begin{figure}[ht]
    \centering
    \begin{minipage}[t]{0.45\textwidth}
        \centering
        \includegraphics[width=\textwidth]{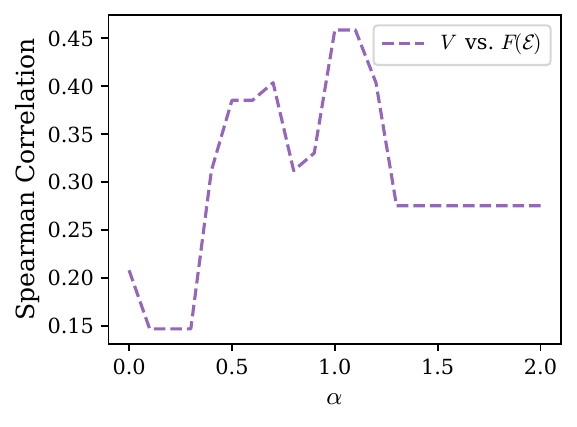}
        \caption{Line plot showing the Spearman correlation between $V$ and $F(\ensemble)$ on GSM8K with different $\alpha$ values.}
        \label{fig:exp:gsm8k_alpha}
    \end{minipage}\hfill
    \begin{minipage}[t]{0.45\textwidth}
        \centering
        \includegraphics[width=\textwidth]{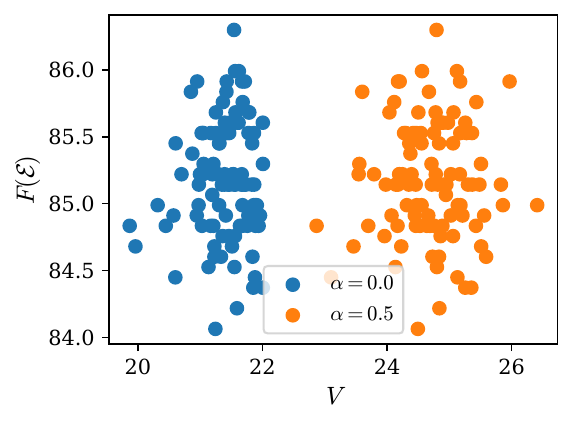}
        \caption{Scatter plot showing the Spearman correlation between $V$ and $F(\ensemble)$ on GSM8K with 2 different $\alpha$ values.}
        \label{fig:exp:gsm8k_scatter}
    \end{minipage}
\end{figure}

We also show that a strong Spearman correlation between $V$ and $F(\ensemble)$ exists for different datasets (e.g., GSM8K). The results presented in \cref{fig:exp:gsm8k_alpha} and \cref{fig:exp:gsm8k_scatter} demonstrate a consistent Spearman correlation between the semantic diversity $V$ and accuracy $F(\ensemble)$ exists. Besides, choosing different fidelity-diversity hyperparameters $\alpha$ may give different results when optimizing for the diversity.

\subsection{Illustration of Semantic Volume}\label{app:sec:illu_logdet}

We illustrate our semantic column metric $V$ here by comparing the semantic volume of (1) a set of 5 prompts generated by just paraphrasing a single original prompt, and (2) a set of 5 prompts randomly selected from the
diverse candidate pool $\mathcal{W}$. \cref{tab:prompts_score} shows how sets of prompts that are qualitatively observed to be more diverse have larger quantitative semantic volume. 

\begin{table}[h]
    \centering
    \resizebox{\linewidth}{!}{%
        \begin{tabular}{p{0.8\linewidth}|c}
            \hline
            \textbf{Set of Prompts} & $V$ \\ \hline
            **Use a Scenario Analysis Approach**: Analyze different scenarios to determine their feasibility and impact. \\
            **Consider Cause and Effect**: Identify potential causes and their effects to understand the question better. \\
            **Use a Benchmarking Approach**: Compare the question to best practices or standards to find the best answer. \\
            **Break Down the Problem**: Divide the question into smaller, manageable parts and tackle each part individually before synthesizing the overall answer. \\
            **Apply Mathematical Logic**: Use mathematical principles and logic to solve the problem, even if it's not a math question. &  -1.24 \\ \hline
            Let's analyze this one step at a time. \\
            Let's break this down step by step. \\
            Let's tackle each part individually. \\
            Let's approach this incrementally. \\
            Let's examine this in a methodical manner. & -1.80 \\
            \hline
        \end{tabular}
    }
    \caption{Example of two sets of prompts with the corresponding diversity score $V$.}
    \label{tab:prompts_score}
\end{table}

\subsection{Generated prompts based on 7 prompts}\label{app:sec:200prompts}

Below in \cref{appx:tab:200prompts} we provide some examples of the generated prompts from GPT-4o based on the 7 prompts.

\begin{table}[h!] 
\centering
\caption{Examples of reasoning prompts generated based on 7 basic prompts.}
\begin{tabular}{p{0.95\columnwidth}}
\hline
\textbf{Prompt} \\
\hline
**Break Down the Problem**: Divide the question into smaller, manageable parts and tackle each part individually before synthesizing the overall answer. \\
\hline
**Apply Mathematical Logic**: Use mathematical principles and logic to solve the problem, even if it's not a math question.\\
\hline
**Use Analogies**: Relate the question to a familiar concept or situation to better understand and solve it. \\
\hline
**Consider the Opposite**: Think about what the answer would be if the opposite were true, to gain a different perspective. \\
\hline
**Consider Cause and Effect**: Identify potential causes and their effects to understand the question better. \\
\hline
\end{tabular}
\label{appx:tab:200prompts}
\end{table}

\end{document}